\documentclass[final]{cvpr}

\usepackage{times}
\usepackage{epsfig}
\usepackage{graphicx}
\usepackage{amsmath}
\usepackage{amssymb}
\usepackage{siunitx}
\usepackage{booktabs}
\RequirePackage{multirow}
\RequirePackage{subdepth}
\makeatletter
\@namedef{ver@everyshi.sty}{}
\makeatother

\RequirePackage{tikz}
\usetikzlibrary{arrows.meta,calc,positioning,shapes.geometric,shapes.symbols,shapes.misc}
\tikzset{%
  >={Latex[width=2mm,length=2mm]},
            base/.style = {rectangle, rounded corners,
                           minimum width=2.25cm, minimum height=0.75cm,
                           text centered},
            input/.style = {trapezium,
                            trapezium left angle=60,
                            trapezium right angle=120,
                            minimum width=2cm, minimum height=0.5cm,
                            text centered},
            line/.style = {draw, -stealth},
            convOp/.style = {base, fill=cyan!30},
            convPost/.style = {base, fill=orange!20},
            inputOp/.style = {input, fill=green!20},
            concatOp/.style = {base, fill=red!20,minimum width=1cm}
}
\newcommand{\FixedLengthArrow}{1,0}

\usepackage{array}
\newcolumntype{M}[1]{>{\centering\arraybackslash}m{#1}}

\usepackage[pagebackref=true,breaklinks=true,colorlinks,bookmarks=false]{hyperref}

\def\cvprPaperID{22} 
\def\confYear{CVPR 2021}

\begin{document}

\title{Temporal Convolution Networks with Positional Encoding\\for Evoked Expression Estimation}

\author{Van Thong Huynh, Guee-Sang Lee, Hyung-Jeong Yang, Soo-Huyng Kim\thanks{Corresponding author}\\
Department of Artificial Intelligence Convergence, Chonnam National University\\
Gwangju 61186, South Korea\\
{\tt\small hvthong.298@outlook.com.vn, \{gslee,hjyang,shkim\}@jnu.ac.kr}
}

\maketitle

\begin{abstract}
   This paper presents an approach for Evoked Expressions from Videos (EEV) challenge, which aims to predict evoked facial expressions from video. We take advantage of pre-trained models on large-scale datasets in computer vision and audio signals to extract the deep representation of timestamps in the video. A temporal convolution network, rather than an RNN like architecture, is used to explore temporal relationships due to its advantage in memory consumption and parallelism. Furthermore, to address the missing annotations of some timestamps, positional encoding is employed to ensure continuity of input data when discarding these timestamps during training. We achieved state-of-the-art results on the EEV challenge with a Pearson correlation coefficient of 0.05477, the first ranked performance in the EEV 2021 challenge.
\end{abstract}

\section{Introduction}
The system with the ability to estimate the emotional impact of video clips would be helpful for a film producer, advertising industry. The evoked expression can change through time together with scene and sound as well as the interaction between them in the video. Over the years, many researches have been conducted mainly focusing on one video type, such as films.

In \cite{yi2019multi}, the authors employed support vector regression with  handcrafted audio features, motion keypoints trajectory, and deep representation of visual and motion features. In the AttendAffectNet approach~\cite{thao2021attendaffectnet}, the authors extracted deep learned features from pre-trained models for audio, visual, and motion information. They employed a self-attention mechanism to take into account and incorporate the relation between modalities. Affect2MM~\cite{mittal2021affect2mm} deployed attention-based and Granger causality to model temporal causality to determine evoked emotional state in movie clips.

In this work, we present our approach for estimating evoked expression from a wide range of video type (e.g. music, entertainment, film, video game, concert). Our method considers the interaction between timestamps with 1D convolutions network and positional encoding to achieve state-of-the-art results on a subset of the EEV dataset~\cite{sun2020eev}.

\section{Our Approach}
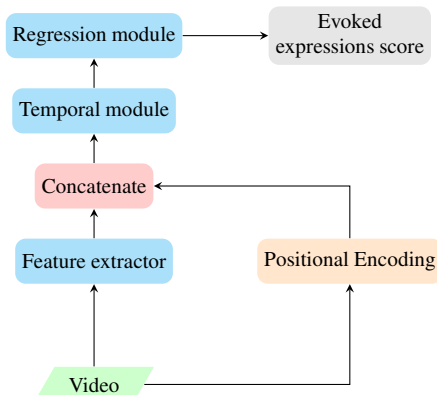
\begin{figure}[!htbp]
    \centering
    \begin{tikzpicture}[node distance=1.25cm, scale=0.8, every node/.style={fill=white, scale=0.8}, align=center]
        \node (inputNode)             [inputOp]              {Video};
        \node (featExtract)     [convOp, above of=inputNode, yshift=0.8cm]          {Feature extractor};
        \node (timeInfo)     [convPost, right of=featExtract,xshift=3cm]          {Positional Encoding};
        \node (concat) [concatOp, above of=featExtract] {Concatenate};
        \node (temporal)     [convOp, above of=concat]          {Temporal module};
        \node (regress)     [convOp, above of=temporal]          {Regression module};
        \node (outputNode) [concatOp, right of=regress, xshift=3cm,fill=gray!20] {Evoked\\expressions score};
        
        \path[-stealth]
            (inputNode)    edge  node[sloped, anchor=center, below, yshift=0.9cm, text width=1.cm, opacity=0, text opacity=1] {Visual or Audio}     (featExtract);
        \draw[-stealth] (inputNode) -| (timeInfo);
        \draw[-stealth] (featExtract) -- (concat);
        \draw[-stealth] (timeInfo) |- (concat);
        \draw[-stealth] (concat) -- (temporal);
        \draw[-stealth] (temporal) -- (regress);
        \draw[-stealth] (regress) -- (outputNode);
    \end{tikzpicture}
    \caption{An overview of our architecture.}
    \label{fig:overview_arch}
\end{figure}

Our work involves two stages to estimate evoked expression for each timestamp in the video: feature extraction, and temporal prediction, as in~\autoref{fig:overview_arch}. In the first stage, we leveraged pre-trained models on the large-scale dataset to extract the audio events or context in the scene. In the following stages, we deployed convolution neural networks and fully connected layers to learn the temporal relationship between timestamps with the help of positional encoding.

\subsection{Feature extraction}
We read the video frames at \SI{6}{\hertz} to extract visual features and resized them to $224\times224$ frames with zero paddings to preserve the aspect ratio.
Then, we fed into the EfficientNet-B0~\cite{tan2019efficientnet}, which is pre-trained for image classification on ImageNet dataset with AutoAugment~\cite{cubuk2018autoaugment} processing, to obtain the features from the last hidden layer (before classification layer), resulting in a 1280-D feature vector for each frame input.

We also used the learnable representation of the audio signal with deep network architecture. We extracted non-overlap audio segments with $1/6$ seconds in length at \SI{16}{\kilo\hertz} to compute log mel-spectrogram representation with 64 mel bands and 96 frames~\cite{hershey2017cnn}. Finally, we used the MobileNet-based distilled version of TRILL (TRIpLet Loss network)~\cite{Shor2020}, pre-trained on AudioSet~\cite{gemmeke2017audio}, to obtain 2048-D feature vector for each audio segment.

\subsection{Temporal and Regression modules}

\begin{figure}[!htbp]
    \centering
    \begin{tikzpicture}[node distance=1.25cm, scale=0.8, every node/.style={fill=white, scale=0.8}, align=center]
        \node (inputNode)             [inputOp]              {Input};
        \node (dilatedConvNode1)     [convOp, above of=inputNode]          {Dilated Casual Conv};
        \node (dilatedConvPost1)     [convPost, above of=dilatedConvNode1]          {WeightNorm,\\ReLU, Dropout};
        \node (dilatedConvNode2)     [convOp, above of=dilatedConvPost1]          {Dilated Casual Conv};
        \node (dilatedConvPost2)     [convPost, above of=dilatedConvNode2]          {WeightNorm,\\ReLU, Dropout};
        \node (conv1x1)				[convOp, right of=dilatedConvPost1, xshift=2cm] {Conv $1\times1$\\(optional)};
        \node (concatOp) [concatOp, right of=dilatedConvPost2, xshift=2cm] {Add};
        
        \draw[-stealth] (inputNode) -- (dilatedConvNode1);
        \draw[-stealth] (dilatedConvNode1)  -- (dilatedConvPost1);
        \draw[-stealth] (dilatedConvPost1) -- (dilatedConvNode2);
        \draw[-stealth] (dilatedConvNode2) -- (dilatedConvPost2);
        \draw[-stealth] (dilatedConvPost2) -- (concatOp);
        \draw[-stealth] (conv1x1) -- (concatOp);
        \draw[-stealth] (inputNode) -| (conv1x1);
        \draw[-stealth] (concatOp) -- ++(\FixedLengthArrow);
    \end{tikzpicture}
    \caption{The details of TCN residual block.}
    \label{fig:tcn_block}
\end{figure}
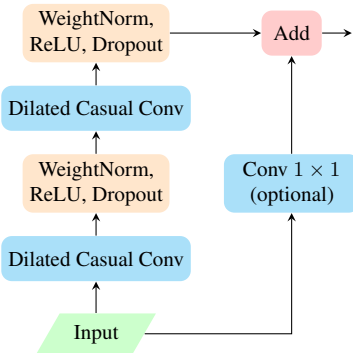

Our temporal module is based on temporal convolution network (TCN)~\cite{bai2018empirical},~\autoref{fig:tcn_block}, a combination of 1-D fully convolution network (FCN), if need to maintain the length of the input, and dilated causal convolutions which achieved long effective history and ensured that there is no information leakage from the future into the past. The dilated casual convolution operation $G$ with a filter $\mathbf{f}$ with size $k$, on element $s$ of the sequence $\mathbf{x}$ can be formulated as 
\begin{equation}
    G(s) = \sum_{i=0}^{k-1} \mathbf{f}_{i}\cdot \mathbf{x}_{s-d\cdot i}, 
\end{equation}
with $d$ is the dilation factor, and the dependence of the output at time $s$ with inputs at current and earlier timesteps is indicated in term of $s-d\cdot i$.
Due to the detection did not occur on some frames of videos, making them unavailable during the annotation process. It makes the discontinuous on frame sequences when using with the temporal module in which we dropped timestamps with no annotation. To ensure continuity, we added positional information into the feature set, which encoded the timestamps for each frame in the input sequence. In the regression module, we used two fully connected layers with ReLU activation between them.

\section{Experimental Results}
    We conducted experiments on the subset\footnote{\url{https://github.com/google-research-datasets/eev}} of Evoked Expressions from Videos (EEV) dataset~\cite{sun2020eev}, a large-scale dataset from diverse videos on YouTube. It consists of 2971, 737, and 1315 videos for training, validation, and testing. Each video is annotated with values in $[0,1]$ at \SI{6}{\hertz} for 15 expressions: amusement, anger, awe, concentration, confusion, contempt, contentment, disappointment, doubt, elation, interest, pain, sadness, surprise, and triumph. The effectiveness $\left(M_{\rho}\right)$ of the model is measured based on the Pearson correlation coefficient ($\rho$) as
    \begin{equation}
        M_{\rho} = \frac{1}{N_{v}}\sum_{j=1}^{N_{v}}\left( \frac{1}{N_{c}}\sum_{i=1}^{N_{c}}\rho_{j,i}\right),
    \end{equation}
    where $N_{c}, N_{v}$ are the number of expression categories and videos, respectively. The $\rho_{j,i}$ indicates $\rho$ score for expression $i^{th}$ in video $j^{th}$ which is formulated as
    \begin{equation}
        \rho_{j,i} = \dfrac{\sum\left[\left(\mathbf{y}_{j,i}-\mu_{\mathbf{y}_{j,i}}\right)\left(\mathbf{\hat{{y}}}_{j,i}-\mu_{\mathbf{\hat{y}}_{j,i}}\right)\right]}{\sqrt{\sum\left(\mathbf{y}_{j,i}-\mu_{\mathbf{y}_{j,i}}\right)^{2}\sum\left(\mathbf{\hat{y}}_{j,i}-\mu_{\mathbf{\hat{y}}_{j,i}}\right)^{2}}},
    \end{equation}
    with $\mathbf{y}_{j,i}$, $\mathbf{\hat{y}}_{j,i}$ are the ground truth and predicted vectors for expression $i^{th}$ in video $j^{th}$. And $\mu_{\mathbf{y}_{j,i}}$, $\mu_{\mathbf{\hat{y}}_{j,i}}$ are the mean of these vectors. The value of $\rho$ falls within the $[-1, 1]$ measures the linear relationship between two vectors with $-1$ indicates a strong negative correlation while $1$ signifies perfect positive correlation, and zero value implying no correlation at all. This also led the value of $M_{\rho}$ ranging from $-1$ to $1$ with similar meaning.
    
    Our feature extractors are based on pre-trained models on TensorFlow Hub, while our temporal and regression modules are deployed with PyTorch Lightning 1.2.6~\cite{falcon2019pytorch} and PyTorch 1.8.1~\cite{paszke2019pytorch}. We trained our models in $20$ epoch using SGD with a learning rate of \num[output-exponent-marker = \text{e}]{5e-3} and a batch size of 32 by using gradient accumulation. With the aim of maximize $M_{\rho}$, we can use the objective function which is formulated as $1-M_{\rho}$, but due to $\rho$ is not defined in case of either $\mathbf{y}_{j,i}$ or $\mathbf{\hat{y}}_{j,i}$ is a constant vector which cause the unstable of gradients during training. Based on that, we used mean squared error as objective function $\left(\mathcal{L}\right)$ during training which is formulated as
    \begin{equation}
        \mathcal{L} = \sum_{j=1}^{N_{b}}\frac{1}{N_{c}}\sum \dfrac{(\mathbf{y}_{j,i}-\mathbf{\hat{y}}_{j,i})^{2}}{L_{j}},
    \end{equation}
    with $N_{b}$ is batch size and $L_{j}$ indicates the length of $\mathbf{y}_{j,i}$ or the number of timestamps in video $j^{th}$. In our training process, batch size of $N_{b}$ corresponding to use $N_{b}$ videos as input.
    \begin{table*}[htbp]
        \centering
        \caption{The comparison of our models in term of $M_{\rho}$ on public and private test set. $d_{p}$ indicates dropout value in TCN residual block.\label{tab:eevResults}}
        \begin{tabular}{p{.07\linewidth}p{.13\linewidth}M{.1\linewidth}M{.2\linewidth}M{.1\linewidth}M{.1\linewidth}} \toprule
            \multicolumn{2}{c}{Feature} & $d_{p}$ & Positional encoding & Public test & Private test \\ \midrule
             Visual & EfficientNet B0 & 0.3 & \checkmark & $0.03440$ & $0.03713$ \\  \midrule
             \multirow{3}{*}{Audio} & MobileNetV2 & $0.3$ & \checkmark & $0.04972$ & $0.04919$ \\
             & MobileNetV2 & $0.3$ &  & $-0.00755$ & $-0.00655$ \\
             & MobileNetV2 & $0.2$ & \checkmark & $0.05060$ & $\mathbf{0.05477}$ \\
             & MobileNetV2 & $0.0$ & \checkmark & $0.04365$ & $0.04371$ \\ \midrule
             \multicolumn{2}{l}{Ensemble} & & & $\mathbf{0.05571}$ & $0.05315$ \\
             \bottomrule
        \end{tabular}
    \end{table*}
    
    ~\autoref{tab:eevResults} shows a comparison between variants of our method. As in the results, the variants with positional encoding outperform the others due to continuity is ensured when feeding input sequence into the temporal module. In our approach, the visual feature is extracted from only one frame for each interval in our approach, while audio representation accounts for the whole interval. It can be the reason why audio outperform visual feature in both public and private test. We also make the ensemble by weighted average as
    \begin{equation}
        S_{\mathrm{ensemble}} = \lambda S_{1} + (1-\lambda)S_{2}
    \end{equation}
    with $\lambda=0.8$, and $S_{1}, S_{2}$ are results of visual and audio features model, including positional encoding and have dropout value ($d_{p}$) of $0.3$, respectively. This lead to an improvement of $8\%$, but still does not outperform the best single audio model with $d_{p}=0.2$ which achieved a Pearson correlation coefficient score of $0.05477$, state-of-the-art on EEV 2021 challenge. Due to the limit on the number of submissions for evaluating the test set, we did not evaluate the visual and best audio model ensemble.
    
\section{Conclusion}
In this work, we presented a method for evoked expression estimation from video clips by incorporating the positional encoding to tackle the missing annotation in each video. Based on that, we achieved state-of-the-art results on the subset of EEV dataset. Our future works mainly focus on improving system performance by exploring the visual information on the whole interval and investigating the interaction between audio and visual models.

\section*{Acknowledgments}
This research was supported by Basic Science Research Program through the National Research Foundation of Korea (NRF) funded by the Ministry of Education (NRF-2018R1D1A3A03000947, NRF-2020R1A4A1019191).

{\small
\bibliographystyle{ieee_fullname}
\bibliography{egbib}
}

\end{document}